# Mixture of ELM based experts with trainable gating network

Laleh Armi[a], Elham Abbasi[a,1], Jamal Zarepour-Ahmadabadi[a]

[a]Department of Computer Science, Yazd University, Yazd, Iran

**Abstract**

Mixture of experts method is a neural network based ensemble learning that has great ability to improve the overall classification accuracy. This method is based on the divide and conquer principle, in which the problem space is divided between several experts by supervisition of gating network.

In this paper, we propose an ensemble learning method based on mixture of experts which is named mixture of ELM based experts with trainable gating network (MEETG) to improve the computing cost and to speed up the learning process of ME. The structure of ME consists of multi layer perceptrons (MLPs) as base experts and gating network, in which gradient-based learning algorithm is applied for training the MLPs which is an iterative and time consuming process. In order to overcome on these problems, we use the advantages of extreme learning machine (ELM) for designing the structure of ME. ELM as a learning algorithm for single hidden-layer feed forward neural networks provides much faster learning process and better generalization ability in comparision with some other traditional learning algorithms. Also, in the proposed method a trainable gating network is applied to aggregate the outputs of the experts dynamically according to the input sample.

Our experimental results and statistical analysis on 11 benchmark datasets confirm that MEETG has an acceptable performance in classification problems. Furthermore, our experimental results show that the proposed approach outperforms the original ELM on prediction stability and classification accuracy.

## 1 Introduction

In the recent decade, noticeable trends have been emerged towards applying computationally efficient machine learning methods. This consideration is because the learning process of neural networks (NNs) based methods is time consuming and inefficient. There are different iterative learning algorithms such as back propagation, Levenberg-Marquardt, heuristics and evolutionary algorithms which have been applied in many researches for training neural network based methods (Slowik and Bialko, 2008; Reynaldi et.al., 2012). Beside the advantages of these methods, they cannot be applied efficiently in the cases that need the fast computations. Because, adjusting the controlling and constructive parameters such as weights, learning rate and momentum need the high computational cost. This problem is more noticeable when the structure of learning system is more complex such as neural network based ensemble systems (Freund et.al.,

[1] Corresponding Author
Department of Mathematics, Yazd University, Yazd, Iran. Tel: :+983531232713,
Email: (l.armi@stu.yazd.ac.ir, e.abbasi@yazd.ac.ir, zarepourjamal@yazd.ac.ir)

1996; Breiman, 1996; Liu and Yao, 1999; Ding et.al., 2014). An ensemble system is a process in which the decisions of multiple experts (classifiers or regressions) are aggregated to improve the overall performance (Zhou and Yu, 2005).

Mixture of experts (ME) as a neural network based ensemble learning method has been considered in many researches (Kheradpisheh et.al., 2014; Masoudnia et.al., 2012; Abbasi et.al., 2016b; Abbasi et.al., 2016a; Peralta and Soto, 2014; Armano and Hatami, 2011; Tang et.al., 2002; Goodband et.al., 2006; Ding et.al., 2014). The structure of ME consists of some experts and a gating network. Based on divide and conquer principle, ME tries to decompose the complex problem into some simple sub problems which this decomposition is managed by gating network. Gating network aggregates the decision of the experts dynamically. In other words, it assigns the weights to the output of the experts by considering the expertise of each expert according to the input sample.

Two important concepts should be considered in designing an ensemble system are the type of base experts and the strategy for combining the outputs of the experts for producing the final decision.

In ME, multilayer perceptrons (MLPs) commonly are applied as the base experts and gating network. Gating network applies a trainable data dependent strategy to aggregate the output of the experts. Despite the strengths of ME, it is associated with some limitations. One common drawback of it is high computation complexity. In ME, gradient descent based algorithms such as back propagation (BP) is applied to train the experts and gating network. Previous researches have shown that the BP method has several limitations such as sensitive to local minima (Cho et.al., 2007), need to set the learning parameters, slow convergence and iterative learning to improve the performance of NNs (Cho et al., 2007; Huang et al., 2006; Zhang et al., 2014).

In 2004, Huang et al. (Huang et al., 2004) found that there is no need to tune the weight between the input layer and hidden layer of feed-forward NNs. Based on this idea, they proposed a novel neural network learning framework which is named extreme learning machine (ELM). This method has been proven to provide an effective, extremely faster learning speed, simple learning algorithm (Huang et al., 2006; Zhang et al., 2016), lack of local minimal and over fitting (Huang et al., 2006), better generalization performance and least human intervention (Ding et al., 2014) compared with some traditional gradient-based learning algorithms (Qiu et al., 2016).

Some researches have been performed to improve the performance of ME. For example, in (Kheradpisheh et al., 2014), a 3-steps ME based ensemble system has been proposed that the experts are trained on different optimal subsets of features. Then, the training of total system has been performed according to standard ME. Also, in (Masoudnia et al., 2012; Abbasi et al., 2016b; Abbasi et al., 2016a) a correlation penalty function is incorporated into the error function of ME. In order to improve the generalization and robustness of ME, in (Abbasi et al., 2016a) a regularization term is embedded into the error function of the proposed method in (Abbasi et al., 2016b). A local feature selection framework based on $l_1$ regularization has been proposed to encourage the experts into different subspaces of the input space (Peralta and Soto, 2014). Also, in (Armano and Hatami, 2011) to specialize the experts in different parts of input space, a mixture of random prototype-based local experts has been proposed. A hybrid ensemble system based on the ME was proposed in which the properties of the ME and boosting were aggregated. In (Tang et al., 2002; Goodband et al., 2006), for improving the input space decomposition, a clustering based algorithm was applied to specialize the experts in different regions of input space. Also, ME and improved versions of it are applied in many applications such as (Ebrahimpour et al., 2008; Armano and Hatami, 2011; Ebrahimpour et al., 2011; Salimi et al., 2012; Lee and Cho, 2014; Mirus

et al., 2019;)

To the best of our knowledge, there are very little researches on the improving the computational complexity of ME. In 2019, Pashaei et al. (Pashaei et al., 2019) have developed an ensemble learning method in which different versions of ELMs such as basic ELM, KELM, OSELM and CELM are applied as base experts for classifying the accident images. Also, they apply three strategies consists of plurality voting method, behavioral knowledge space (BKS) and decision making patterns for aggregating the outputs of experts.

In continuation of these works, in this paper, we develop an ensemble learning method based on ME which is named mixture of ELM based experts with trainable gating network (MEETG) to improve the computing cost of ME. We use the advantages of ELM for building the base experts and gating network. In BP based ME, the high computational complexity is introduced by tuning the weights of the experts and gating network, iteratively. In the proposed method, the learning process is not iterative. It consists of adjusting only the weights between hidden layer and output layer in one pass. The experimental results show that MEETG performs better in comparison with standard ME and state-of-the-art method.

The rest of this paper is as follows: in the Section 2, we will overview the preliminaries concepts. In Section 3, we will describe the proposed method. Experimental results are discussed in Section 4 and finally, a general conclusion is provided in Section 5.

## 2 Preliminaries

In this section, we briefly review extreme learning machine and mixture of expert to provide necessary backgrounds for the proposed method.

### 2.1 Extreme learning machine

Extreme learning machine (ELM) as an emerging learning algorithm for training single-hidden-layer feed-forward neural networks (SLFNs) has been proposed by Huang et al. (Huang et al., 2004). Compared to conventional neural network learning algorithms such as back propagation (BP) algorithm, the training speed of ELM is extremely fast. This usefulness is because of that it's learning process performs only in one pass and it is not iterative (Qiu et al., 2016). Also, the computational cost of ELM is much low. In ELM, the weights between input layer and hidden layer are set randomly and remain constant during training. On the contrary, the weights between hidden layer and output layer are determined analytically by applying a linear least squares method (Janakiraman et al., 2016). Some other attractive features of ELM include good generalization performance, high classification accuracy and few adjustable parameters, avoiding time-costing iterative learning process and also avoiding falling into local minima (Huang et al., 2006).

Suppose $D = \{(x_i, y_i)_{i=1,2,\cdots,N_{train}}. x_i \epsilon \Re^d\}$ denote the training dataset which consists of $N_{train}$ training samples with $d$ input features, and $m$ classes. $x_i = [x_{i1}, x_{i2}, \cdots, x_{id}]^T$ and $y_i = [y_{i1}, y_{i2}, \cdots, y_{im}]^T$ are input data and desired output, respectively. The number of hidden layer neurons in ELM is $L$ and activation function is as follows:

$$G(w_j, b_j, x_i) = \frac{1}{1+exp(-(w_j.x_i+b_j))} \quad (1)$$

where $w_j = [w_{j1}, w_{j2}, \cdots, w_{jd}]$ and $b_j$ are the input weight and bias between input layer

and $j^{th}$ hidden neuron which is set randomly. So, ELM model can be described as follows:

$$H\beta = Y \qquad (2)$$

where

$$H = \begin{bmatrix} G(w_1, b_1, x_1) & \cdots & G(w_L, b_L, x_1) \\ \vdots & \ddots & \vdots \\ G(w_1, b_1, x_{N_{train}}) & \cdots & G(w_L, b_L, x_{N_{train}}) \end{bmatrix}_{N_{train} \times L} \qquad (3)$$

$$\beta = \begin{bmatrix} \beta_{11}, \beta_{12}, \cdots, \beta_{1m} \\ \vdots \\ \beta_{j1}, \beta_{j2}, \cdots, \beta_{jm} \\ \vdots \\ \beta_{L1}, \beta_{L2}, \cdots, \beta_{Lm} \end{bmatrix}_{L \times m} \quad and \quad Y = \begin{bmatrix} y_{11} & \cdots & y_{1m} \\ \vdots & & \\ y_{N_{train}1} & \cdots & y_{N_{train}m} \end{bmatrix}_{N_{train} \times m} \qquad (4)$$

According to Eq. 3, $H$ is the hidden layer output matrix. The $j^{th}$ column of $H$ is $j^{th}$ hidden neuron's output vector with respect to $x_1, x_2, \cdots, x_{N_{train}}$. Also, $\beta_j = [\beta_{j1}, \beta_{j2}, \cdots, \beta_{jm}]$ is output weight between $j^{th}$ hidden neuron and output layer and it is calculated as the following:

$$\beta = H^\dagger . Y \qquad (5)$$

where $H^\dagger = (HH^T)^{-1}H$ is the Moore-Penrose generalized inverse of matrix $H$ and $Y$ is the training data target matrix (Huang et al., 2006). Finally, output of ELM is calculated according to Eq. 6.

$$f(SampleTest_i) = \sum_{j=1}^{L} \beta_j G(w_j, b_j, SampleTest_i) \quad i = 1, \ldots, N_{test} \qquad (6)$$

## 2.2 Mixture of experts

Mixture of experts (ME) is one of the most popular ensemble learning method which is introduced by Jacobs et al. (Jacobs et al., 1991). The modular architecture of ME consists of some experts and a gating network which learns the input space decomposition and also aggregates the outputs of experts dynamically according to the expertise of each expert in created subspaces. Let $N$ be the number of experts which is denoted by $j = 1, 2, \ldots, N$. The output of each expert is weighted by the gating network as follows:

$$g_i = \frac{\exp(O_{g_i})}{\sum_{j=1}^{N}(O_{g_j})} \qquad (7)$$

where $O_{g_i}$ is the $i^{th}$ output of the gating network and the gating network produces $g_i$ for each expert with respect to the each input vector. The $g_i$ can be interpreted as estimation of selecting the output from $i^{th}$ expert by the gating network.

According to Eq. 7, the softmax function is applied as the gating network which satisfies $g_i \geq 0$ and $\sum g_i = 1$. The final output of ME as an aggregation of the gating network output and the expert output is calculated as follows:

$$O_{ens} = \sum_{i=1}^{N} O_i . g_i \qquad (8)$$

Figure.1 illustrates the structure of the ME.

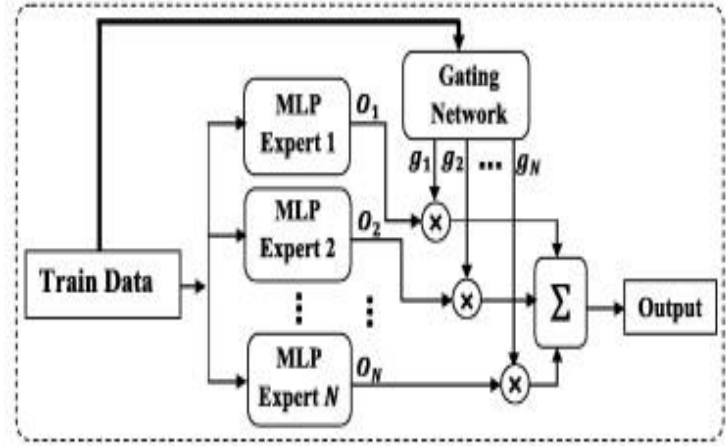

Figure 1. Mixture of experts block diagram (Kheradpisheh et al., 2014)

The error function of the gating network that is given in (Masoudnia and Ebrahimpour, 2014) is calculated as Eq. 9.

$$E_G = \frac{1}{2}\|h - O_g\|^2 \tag{9}$$

where $h = \{h_i\}_{i=1}^{N}$ is a posterior probability that estimates to what extent expert $i$ can produce the desired output $y$ and $O_g$ is the output of gating network. Also, $h_i$ is calculated as the following:

$$h_i = \frac{g_i \exp(\frac{-1}{2}\|y - O_i\|^2)}{\sum_j g_i \exp(\frac{-1}{2}\|y - O_j\|^2)} \tag{10}$$

## 3  Mixture of ELM based experts with trainable gating network (MEETG)

In this section, we introduce our proposed approach, mixture of ELM based experts with trainable gating network (MEETG) that it takes advantages of the ELM in designing the structure of ME. As we mentioned in the previous section, mixture of experts is one of the most popular ensemble learning based methods which can improve the accuracy for the complex problems. ME has some considerable features which make it different from some other ensemble learning methods. One of them is the great potential of it to localize the experts in different parts of input space.The other one is the ability of ME in aggregating the outputs of experts. Unlike the most ensemble learning methods that assign fixed weights to the output of base classifiers, in ME, the output of each expert is weighted dynamically by the gating network according to the input data and local expertise of the experts in different areas of problem space.

Despite the strengths of the ME, it is associated with some limitations. Basically, in ME, MLPs are as base experts and gating network. The training process of MLPs by applying gradient descent based algorithms is time-consuming and may easily converge to local minimums. Also, in ME, neural network based models should be applied as experts.

On the other hand, ELM as a learning algorithm for single hidden-layer feed forward neural networks compared with gradient descent based learning algorithms such as back propagation (BP), has some advantages such as extremely high training speed, good generalization ability, not needing to set the parameters such as stopping criteria or learning rate, partially overcoming the problem of over fitting and local minimal (Zhang et al., 2014).

According to the advantages of ELM and ME and also to cover the limitation of ME which being specific to NN, ELM can be applied for designing the structure of ME.

So, in this paper, the properties of ELM and ME are integrated in which the ELMs are applied as base experts and gating network, entitled as, mixture of ELM based experts with trainable gating network (MEETG) to overcome on the limitations of gradient descent based learning algorithms and to improve the computing cost of ME.

Figure. 2 shows the structure of MEETG. As shown in Figure. 2, the input samples $X_{N_{train} \times d}$ that $d$ is the dimension of input are sent to the experts and gating network simultaneously. The output layer of each expert consists of $m$ neurons corresponding to the target vector $[y_1, \cdots, y_m] \epsilon\ Y_{N_{train} \times m}$.

The weights between input layer and hidden layer of each expert are assigned randomly. The sigmoid function is applied as the activation function of the neurons in the hidden layer. Also, the weights between hidden layer and output layer are calculated according to Eq.5. In MEETG, the output of each expert is weighted by the gating network according to Eq. 8. Also, the number of output neurons in gating network for a system with $k$ experts is equal to $k$. According to Eq. 10, $h$ as a tareget vector of gating network is given as the following:

$$h = [\frac{\exp\|y - O^{Expert\ 1}\|^2}{\sum_{i=1}^{k}\exp\|y - O^{Expert\ i}\|^2}, \cdots, \frac{\exp\|y - O^{Expert\ k}\|^2}{\sum_{i=1}^{k}\exp\|y - O^{Expert\ i}\|^2}] \qquad (11)$$

where $y$ is the target output and $O^{Expert\ i}$ is the output of expert $i$. $H$ as the target matrix of the gating network for $N_{train}$ training sapmles and $k$ experts is given as the following:

$$H = \begin{bmatrix} \frac{\exp\|y_1 - O_1^{Expert\ 1}\|^2}{\sum_{i=1}^{k}\exp\|y_1 - O_1^{Expert\ i}\|^2} & \cdots & \frac{\exp\|y_1 - O_1^{Expert\ k}\|^2}{\sum_{i=1}^{k}\exp\|y_1 - O_1^{Expert\ i}\|^2} \\ \vdots & \ddots & \vdots \\ \frac{\exp\|y_{N_{train}} - O_{N_{train}}^{Expert\ 1}\|^2}{\sum_{i=1}^{k}\exp\|y_{N_{train}} - O_{N_{train}}^{Expert\ i}\|^2} & \cdots & \frac{\exp\|y_{N_{train}} - O_{N_{train}}^{Expert\ k}\|^2}{\sum_{i=1}^{k}\exp\|y_{N_{train}} - O_{N_{train}}^{Expert\ i}\|^2} \end{bmatrix}_{N_{train} \times k} \qquad (12)$$

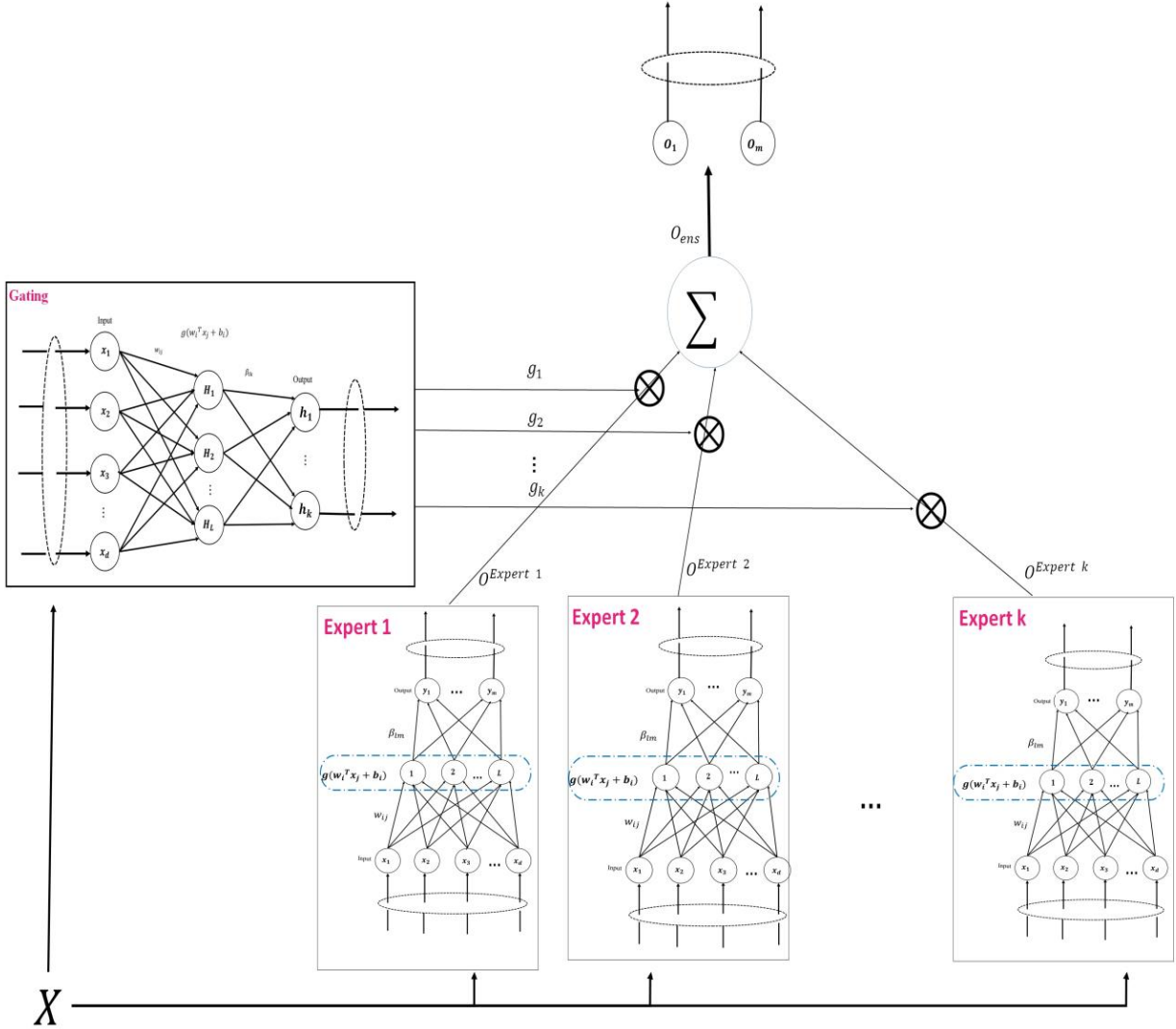

Figure 2. Block diagram of MEETG

Similar to the experts, ELM is applied as the gating network and its weight is adjusted according to Eq. 5. Finally, to produce the final output, the output of each expert is aggregated with the corresponding output of gating network by Eq. 8. For final decision making, the maximum possibility is considered as follows:

$$c = argmax_{j=1}^{m} O_{ens} \tag{13}$$

where $O_{ens} = [O_1, O_2, ..., O_m]$ is the output vector of MEETG corresponding to *m* classes for each test sample. The training mechanism of MEETG algorithm is presented in Algorithm 1.

| **Algorithm 1:** Training mechanism of MEETG algorithm | | |
|---|---|---|
| **Input:** | $D = \{(x_i, y_i)_{i=1,2,\cdots,N_{train}}. x_i \epsilon \Re^d$ and $y_i \epsilon \Re^m\}$: set of training samples, $L$: number of hidden layer neurons, $k$: number of experts, $X_{test}$: test sample | |
| **Output:** | c: output class | |
| **Train step:** | Step1) | Assign input weight $w = [w_1, w_2, \ldots, w_L]$ and bias $B = [b_1, b_2, \ldots, b_L]$ for each expert and gating network randomly. |
| | Step 2) | Calculate the hidden layer output matrix for each expert and gating network according to Eq.3. |
| | Step 3) | Calculate the desired output of the gating according to Eq.12. |
| | Step 4) | Calculate the weights between hidden and output layers for each expert according to Eq.5. |
| **Test step:** | Step1) | Calculate the output of the experts and gating network according to Eq.6. |
| | Step2) | The output of each expert is weighted by the gating network by Eq.7. |
| | Step3) | Calculate the final output according to Eq.8. |
| | Step4) | Determine the final class according to Eq. 13. |

In the next section, the performance of the proposed method is compared with other techniques applying eleven benchmark datasets.

## 4 Experimental results

To illustrate the performance of the proposed method, we report the experimental results of our method on eleven benchmark datasets which they have been adopted from the UCI Machine Learning Repository (Asuncion and Newman, 2007). The breif description of the datasets are shown in Table 1.

Table 1. The basic information of the eleven classification benchmarks

| Data set | size | ♯class | ♯Attribute |
|---|---|---|---|
| Balance | 625 | 3 | 4 |
| Breast | 277 | 2 | 9 |
| Glass | 217 | 7 | 10 |
| Heart | 270 | 2 | 13 |
| Iris | 150 | 3 | 4 |
| Lymphography | 148 | 4 | 8 |
| Pen-digit | 10992 | 10 | 16 |
| Sat-image | 6435 | 6 | 4 |
| Thyroid | 215 | 3 | 5 |
| Twonorm | 7400 | 2 | 20 |
| Yeast | 1484 | 10 | 8 |

In two next sub sections, we compare our proposed method with a single ELM and with some other methods such as Bagging, Boosting, Adaboost and ME.

## 4.1 Comparison between the performance of MEETG and single ELM

All experiments are performed with MATLAB 2018a under the Windows 10 Enterprise operating system on a Core i5 and 2.40 GHz computer with an 12 GB RAM.

Comparison between the performance of MEETG and single ELM on several classification problems are performed (Asuncion and Newman, 2007) in these experiments. Also, k-fold cross-validation with k = 5 and k=10 is used to measure the performance of the methods. A comparative analysis of the accuracy between MEETG and standard ELM is presented in Figure. 3.

The accuracy is improved more than 10% for some datasets such as Yeast, Sat-Image, Thyroid, Lymph, and Glass. Also, the accuracy of Twonorm dataset and Pen-digit dataset is about 0.68% and 0.4% better than ELM, respectively. The results indicate ensemble learning based proposed method performs better than single ELM.

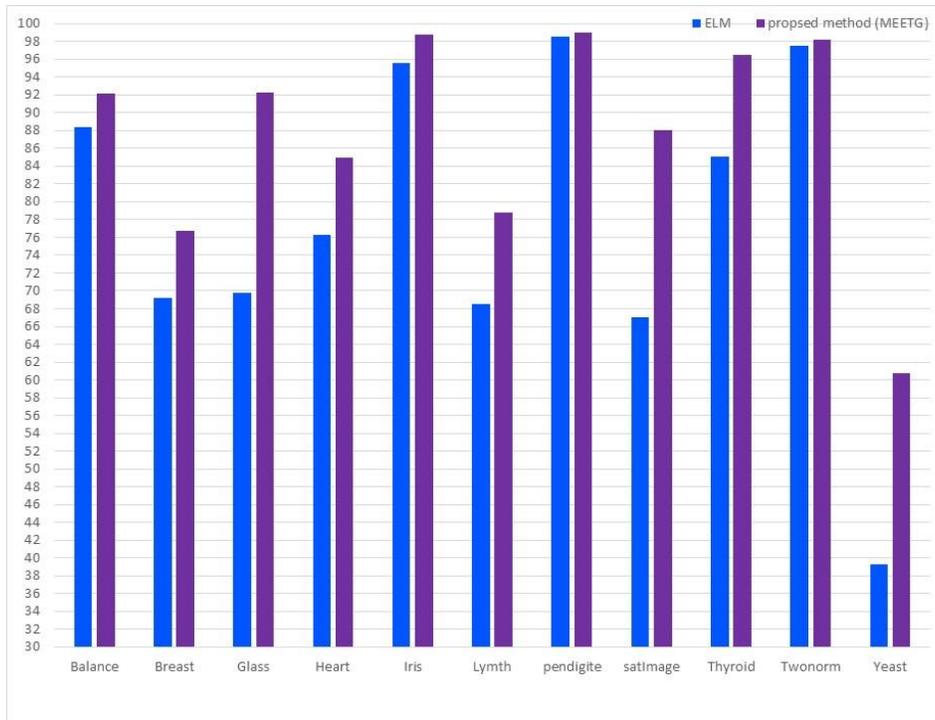

Figure 3. Comparison between the proposed method and single ELM

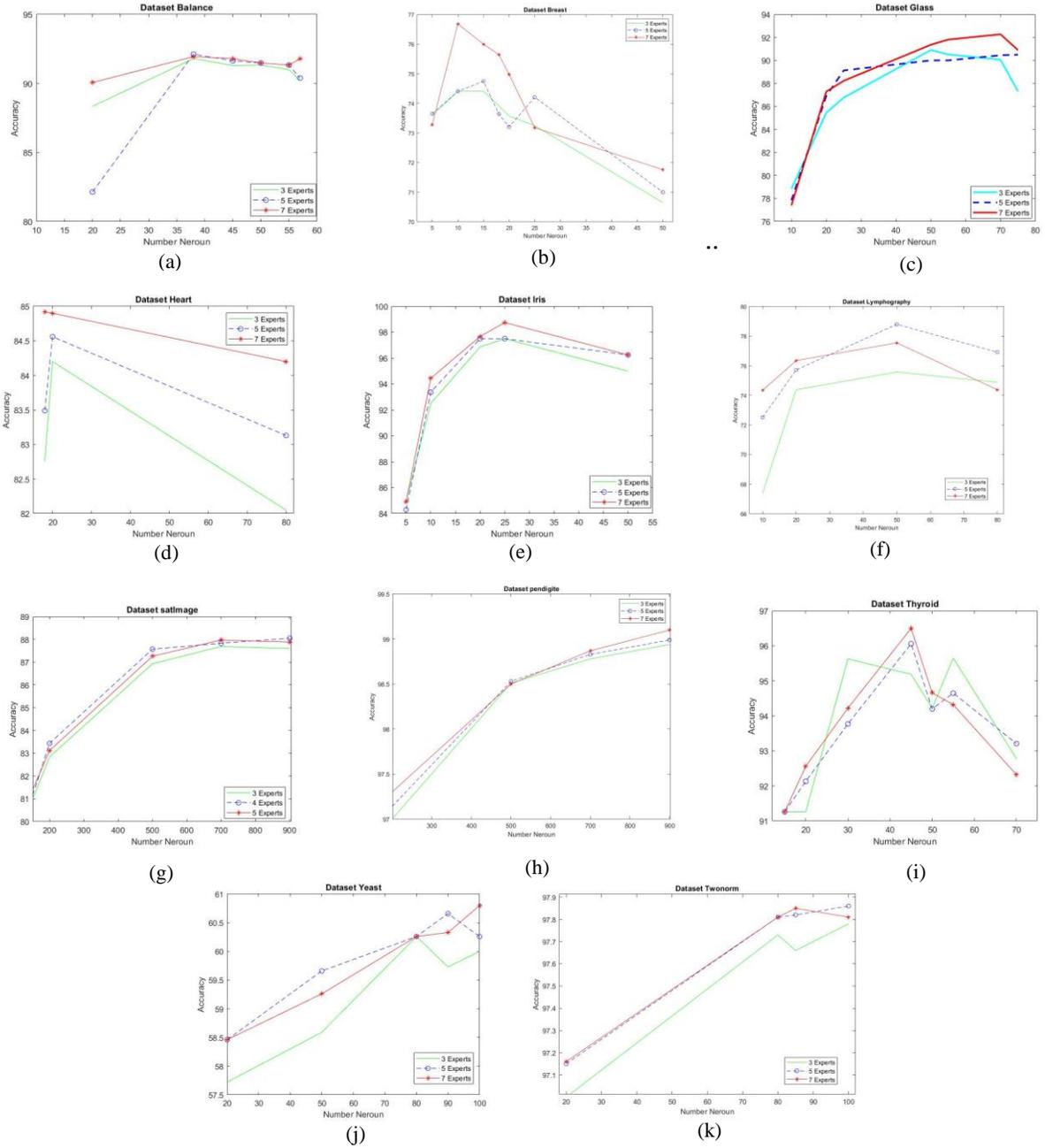

Figure 4. The accuracy for three, five, and seven experts with different number of hidden layer neurons on (a) Balance dataset,(b) Breast dataset, (c) Glass dataset, (d) Heart dataset, (e) Iris dataset, (f) Lymphography dataset, (g) Prndigit dataset, (h) SatImage dataset, (i) Thyroid dataset, (j) Yeast dataset, (k) Twonorm dataset.

The effect of changing the number of experts and the number of hidden layer neurons is depicted in Figure 4. Different number of hidden neurons and different ensemble size including three, five, and seven experts are analyzed based on 10-fold cross validation. According to Figure. 4 (c),(e), (g),(h) and (k) ,increasing the number of hidden neurons with seven experts improves the accuracy of the MEETG method.

In Figure 4 (b), the best accuracy is 76.69% which is obtained using seven experts and 10 hidden neurons. In Balance dataset, five experts and 38 hidden neurons present the best accuracy.

As shown in Figure 4 (d), regardless of the number of experts, 18 hidden neurons give the best accuracy. For the Lymph dataset, Figure 4 (f), the best result is achieved with 55 hidden neurons.

## 4.2 Comparison between the perormance of MEETG and state-of-the-are methods

In this subsection the performance of the proposed method has been compared with some of state-of-the-are-methods. The best performance of the Bagging, Boosting & Adaboos and ME is adopted from (Abbasi et al., 2016b; Abbasi et al., 2016a). The results are summarized in Table 2. The last two columns in Table 2. represent the performance of our proposed method with 5-fold and 10-fold cross validation.

Table 2. The classification test error rates (%) of different ensemble methods for eleven classification benchmark datasets

| Data set | methods base | | | | |
| --- | --- | --- | --- | --- | --- |
| | Bagging | Boosting & Adaboost | ME | MEETG 5-fold | MEETG 10-fold |
| Balance | 7 | 6.8 | 6.7 | 9.11 | 7.88 |
| Breast | 27.3 | 30.4 | 25.45 | 24.69 | 23.31 |
| Glass | 29.9 | 32 | 30.8 | 9.89 | 7.27 |
| Heart | 17.2 | 20.3 | 15.57 | 16.87 | 15.08 |
| Iris | 4.4 | 5.6 | 4.67 | 0.44 | 1.23 |
| Lymph | 18.4 | 23.33 | 16.10 | 17.6 | 21.6 |
| Pen-digit | 7 | 6.2 | 6.4 | 1.07 | 0.99 |
| Sat-imag | 17 | 16.1 | 16.2 | 13.55 | 11.95 |
| Thyroid | 4.4 | 4.4 | 5.58 | 4.37 | 3.5 |
| Twonorm | 2.8 | 3 | 2.69 | 1.82 | 2.14 |
| Yeast | 42.6 | 42.1 | 40.5 | 41.32 | 39.2 |

Table 3. shows a comparison between the performance of proposed method and other ensemble learning methods. In this table, we compare our proposed method with MR-FI-ELM (Zhai et al., 2018), RTQRT-ME (Abbasi et al., 2016b), DCE-CC (Li et al., 2013), (Ebrahimpour et al., 2013) and Vote B&B&R&R (Kotsiantis, 2011).

In (Ebrahimpour et al., 2013), an improved version of NCL method in which the gating network capability, as the combining part of ME method, is used to combine the base NNs in the NCL ensemble method. In (Zhai et al., 2018), fuzzy integral has been applied to integrate the trained basic classifiers. Extreme learning machine algorithm was used as base classifier in the ensemble learning method. In (Kotsiantis, 2011), bagging, boosting, rotation forest and random sub space were combined. In each method, 6 sub-classifiers have been applied and a voting strategy has been used for the final decision. Also, decision tree has been used as base classifier in the ensemble.

Table 3. Comparison the proposed method with other ensemble learning methods

| Data set | Method | Accuracy |
|---|---|---|
| Balance | (Li et al., 2013) | 75.18 |
| | (Abbasi et al., 2016b) | 92.96 |
| | **MEETG** | 92.12 |
| Breast | (Abbasi et al., 2016b) | 75.74 |
| | (Kotsiantis, 2011) | 72.07 |
| | **MEETG** | **76.69** |
| Glass | (Li et al., 2013) | 72.48 |
| | (Abbasi et al., 2016b) | 72.87 |
| | (Ebrahimpour et al., 2013) | 69.4 |
| | **MEETG** | **92.27** |
| Heart | (Li et al., 2013) | 80 |
| | (Abbasi et al., 2016b) | 84.41 |
| | (Kotsiantis, 2011) | 81.74 |
| | **MEETG** | **84.92** |
| Iris | (Kotsiantis, 2011) | 94.87 |
| | (Zhai et al., 2018) | 96.58 |
| | **MEETG** | **98.85** |
| Pen-digit | (Abbasi et al., 2016b) | 97.13 |
| | (Ebrahimpour et al., 2013) | 93.2 |
| | **MEETG** | **99.10** |
| sat-Image | (Abbasi et al., 2016b) | 87.44 |
| | (Ebrahimpour et al., 2013) | 85.6 |
| | **MEETG** | **88.50** |
| Thyroid | (Li et al., 2013) | 96.39 |
| | **MEETG** | **96.50** |
| Yeast | (Li et al., 2013) | 67.78 |
| | (Abbasi et al., 2016b) | 62.69 |
| | **MEETG** | **60.80** |

According to Table 3., the performance of MEETG is 1.97% and 5.9% better than (Abbasi et al., 2016b) and (Ebrahimpour et al., 2013) for Pen-digit dtaset. Also, the performance of MEETG is 19.79% better than (Li et al., 2013) and (Abbasi et al., 2016b) for Glass dataset. The performance of (Abbasi et al., 2016b) is about 0.84% better than MEETG for Balance dataset. But MEETG performs 16.94% better than (Li et al., 2013) on the Balance dataset.

## 4.3 Evaluation of MEETG

Precision and recall are applied in addition to overall classification accuracy (Nekooeimehr and Lai-Yuen, 2016) to evaluate the performance of the proposed method according to Eq. 14.

$$Precision = \frac{TP}{TP+FP} \qquad Recall = \frac{TP}{TP+FN} \qquad (14)$$

where precision is defined as the number of true positives ($TP$) over the number of true positives plus the number of false positives ($FP$). Recall is defined as the number of true positives ($TP$) over the number of true positives plus the number of false negatives ($FN$). The results of applying these metrics are presented in Table 4.

Table 4. Precision and recall metrics of the proposed method and (Abbasi et al., 2016b) on eleven datasets

| Data set | Precision (MEETG) | Precision (Abbasi et al., 2016b) | Recall (MEETG) | Recall (Abbasi et al., 2016b) |
|---|---|---|---|---|
| Balance | 0.9302 | 0.91 | 0.8797 | 0.81 |
| Breast | 0.6593 | - | 0.7208 | |
| Glass | 0.8726 | 0.74 | 0.8569 | 0.67 |
| Heart | 0.8433 | - | 0.8575 | |
| Iris | 0.9822 | - | 0.9896 | |
| Lymph | 0.7603 | - | 0.7404 | |
| Pen-digit | 0.9901 | 0.97 | 0.9901 | 0.97 |
| Sat-Image | 0.8516 | 0.85 | 0.8418 | 0.82 |
| Thyroid | 0.9228 | - | 0.09845 | |
| Twonorm | 0.9818 | - | 0.9819 | |
| Yeast | 6943 | 0.64 | 0.7214 | 0.62 |

## 4.4 Computation time

As claimed in this research, the computational complexity of our proposed approach is not high. The run-time of each classifier depends on the number of parameters and the nature of the problem such as the number of input samples, the dimension of input samples and the number of classes involved in the problem. In order to investigate the computational complexity of the proposed method, two experiments are evaluated as follows:
- Total number of required operations
- Real running time

The run-time performance of MEETG can be decomposed as three main components: 1) expert networks, 2) gating network, and 3) combination phase. As mentioned previously, ELM network is applied as the base experts and gating network in MEETG. The number of computational operators is considered in the training phase. Totally, the number of computational operators can be given as the following:

$$T_{MEETG} = \sum_{k=1}^{N_{Expert\ k}} T_k + T_g + T_c \qquad (17)$$

where $T_k$, $T_g$ and $T_c$ are total number of operations for each expert, gating network and aggerigation phase, respectively.

Suppose $C$ is the number of output classes, $d$ is the dimension of the feature vector and l is the number hidden layer neurons. Also, $N_{wh} = d \times l$ is the number of weights between input layer and hidden layer and $N_{wo} = l \times C$ is the number of weights between hidden layer and output layer. The total number of operations for each expert corresponding to $N_{sampelTrain}$ samples is as follows:

$$T_{expert} = N_{sampelTrain} \times d \times l) \qquad (18)$$

For gating network, parameters of the gating network, such as the number of neurons in the hidden and in the output layer, are usually different. Given the equation below, the symbol $N_g$ represents number of output neurons

$$T_{gating} = N_{sampelTrain} \times (d \times N_{w0} \times N_g) \quad (19)$$

Also, the number of operations required to calculate the weighted output of experts is:

$$T_{combin} = C \times N_g \quad (20)$$

Also, the running time of the proposed method are reported in Table 5. In this experiment, we used 10-fold cross validation to evaluate the accuracy of the classifications.

Table 5. The average of running time of MEETG in seconds.

|  | Dataset | | | | | |
|---|---|---|---|---|---|---|
|  | Balance | Breast | Glass | Heart | Iris | Lymph |
| Running time (MEETG): | 0.2369 | 0.1395 | 0.1133 | 0.1461 | 0.13916 | 0.1455 |
|  | Dataset | | | | | |
|  | Pen-digit | sat-image | Thyroid | Twonorm | Yeast | |
| Running time (MEETG): | 16.932 | 45.6969 | 0.20114 | 2.4628 | 1.1532 | |

## 5 Conclusion

In this paper, we propose a novel ME based ensemble learning approach named as mixture of ELM based experts with trainable gating network (MEETG). Despite the strengths of ME, it is associated with some limitations. One common drawback of it is high computation complexity. The base experts at conventional ME are MLPs, which are known for their slow training process and gradient descent based algorithms such as back propagation (BP) is applied to train the experts and gating network. Since, ELM has different advantages such as high training speed and good generalization ability, in this paper, the advantages of ELM is taken in designing the structure of ME. The advantages of the proposed MEETG algorithm include non-iterative learning and high training speed, not needing manual intervention like setting such as stop criterion or learning rate and good performance. Also, a trainable gating network is applied to aggregate the output of the experts according to their expertise in different regions of input space.

Practical results show that our proposed method is acceptable in most cases.

## References

Abbasi, E., Shiri, M. E., and Ghatee, M. (2016a). A regularized root–quartic mixture of experts for complex classification problems. *Knowledge-Based Systems*, 110, 98–109.

Abbasi, E., Shiri, M. E., and Ghatee, M. (2016b). Root-quatric mixture of experts for complex classification problems. *Expert Systems with Applications*, 53, 192–203.

Armano, G. and Hatami, N. (2011). An improved mixture of experts model: Divide and conquer using random prototypes. In *Ensembles in Machine Learning Applications*, pages 217–231. Springer.

Asuncion, A. and Newman, D. (2007). UCI machine learning. *NeuralComputation.* C. U. o. C. Irvine, School of Information and Computer Science http://www.ics.uci.edu/~mlearn/ML Repository. html.

Breiman, L. (1996). Bagging predictors. *Machine learning*, 24(2), 123–140.

Cho, J.-H., Lee, D.-J., and Chun, M.-G. (2007). Parameter optimization of extreme learning machine using bacterial foraging algorithm. *Journal of Korean Institute of Intelligent Systems*, 17(6), 807–812.

Ding, S., Xu, X., and Nie, R. (2014). Extreme learning machine and its applications. *Neural Computing and Applications*, 25(3-4), 549–556.

Ebrahimpour, R., Kabir, E., Esteky, H., & Yousefi, M. R. (2008). View-independent face recognition with mixture of experts. *Neurocomputing*, 71(4-6), 1103-1107.

Ebrahimpour, R., Arani, S. A. A. A., and Masoudnia, S. (2013). Improving combination method of ncl experts using gating network. *Neural Computing and Applications*, 22(1), 95–101.

Ebrahimpour, R., Sarhangi, S., and Sharifizadeh, F. (2011). Mixture of experts for persian handwritten word recognition.

Freund, Y., Schapire, R. E., et al. (1996). Experiments with a new boosting algorithm. In *icml*, volume 96, pages 148–156. Citeseer.

Goodband, J., Haas, O. C., and Mills, J. A. (2006). A mixture of experts committee machine to design compensators for intensity modulated radiation therapy. *Pattern recognition*, 39(9), 1704–1714.

Huang, G.-B., Zhu, Q.-Y., and Siew, C.-K. (2006). Extreme learning machine: Theory and applications. *Neurocomputing*, 70(1), 489 – 501.

Huang, G.-B., Zhu, Q.-Y., and Siew, C. (2004). Extreme learning machine: A new learning scheme of feedforward neural networks. 2(2), 985 – 990.

Jacobs, R. A., Jordan, M. I., Nowlan, S. J., Hinton, G. E., et al. (1991). Adaptive mixtures of local experts. *Neural computation*, 3(1), 79–87.

Janakiraman, V. M., Nguyen, X., and Assanis, D. (2016). Stochastic gradient based extreme learning machines for stable online learning of advanced combustion engines. *Neurocomputing*, 177, 304–316.

Kheradpisheh, S. R., Sharifizadeh, F., Nowzari-Dalini, A., Ganjtabesh, M., and Ebrahimpour, R. (2014). Mixture of feature specified experts. *Information Fusion*, 20, 242–251.

Kotsiantis, S. (2011). Combining bagging, boosting, rotation forest and random subspace methods. *Artificial Intelligence Review*, 35(3), 223–240.

Lee, Y. S., and Cho, S. B. (2014). Activity recognition with android phone using mixture-of-experts co-trained with labeled and unlabeled data. *Neurocomputing*, 126, 106-115.

Li, L., Zou, B., Hu, Q., Wu, X., and Yu, D. (2013). Dynamic classifier ensemble using classification confidence. *Neurocomputing*, 99, 581–591.

Liu, Y. and Yao, X. (1999). Ensemble learning via negative correlation. *Neural networks*, 12(10), 1399–1404.

Masoudnia, S., Ebrahimpour, R., and Arani, S. A. A. A. (2012). Incorporation of a regularization term to control negative correlation in mixture of experts. *Neural processing letters*, 36(1), 31–47.

Masoudnia, S. and Ebrahimpour, R. (2014). Mixture of experts: a literature survey. *Artificial Intelligence Review*, 42(2), 275–293.

Mirus, F., Stewart, T. C., Eliasmith, C., & Conradt, J. (2019). A Mixture-of-Experts Model for Vehicle Prediction Using an Online Learning Approach. In *International Conference on Artificial Neural Networks*, pages 456-471.

Nekooeimehr, I., & Lai-Yuen, S. K. (2015). Adaptive semi-unsupervised weighted oversampling (A-SUWO) for imbalanced datasets. *Expert Systems with Applications*, 46, 405-416.

Pashaei, A., Ghatee, M., and Sajedi, H. (2019). Convolution neural network joint with mixture of extreme learning machines for feature extraction and classification of accident images. *Journal of Real-Time Image Processing*. 1-16.

Peralta, B. and Soto, A. (2014). Embedded local feature selection within mixture of experts. *Information Sciences*, 269, 176–187.

Qiu, J., Wu, Q., Ding, G., Xu, Y., and Feng, S. (2016). A survey of machine learning for big data processing. *EURASIP Journal on Advances in Signal Processing*, 2016(1), 67.

Reynaldi, A., Lukas, S., & Margaretha, H. (2012, November). Backpropagation and Levenberg-Marquardt algorithm for training finite element neural network. In 2012 *Sixth UKSim/AMSS European Symposium on Computer Modeling and Simulation,* pages 89-94. IEEE.

Salimi, H., Giveki, D., Soltanshahi, M. A., and Hatami, J. (2012). Extended mixture of mlp experts by hybrid of conjugate gradient method and modified cuckoo search. *International Journal of Artificial Intelligence & Applications (IJAIA),* 3(1), 1-13.

Slowik, A. and Bialko, M., (2008, May). Training of artificial neural networks using differential evolution algorithm. In *2008 conference on human system interactions*, pages 60-65. IEEE.

Tang, B., Heywood, M. I., and Shepherd, M. (2002, May). Input partitioning to mixture of experts. In *Proceedings of the 2002 International Joint Conference on Neural Networks. IJCNN'02 (Cat. No. 02CH37290)*, (1), pages 227–232. IEEE.

Zhai, J., Zang, L., and Zhou, Z. (2018). Ensemble dropout extreme learning machine via fuzzy integral for data classification. *Neurocomputing*, 275, 1043 – 1052.

Zhang, H.-G., Zhang, S., and Yin, Y.-X. (2014). A novel improved elm algorithm for a real industrial application. *Mathematical Problems in Engineering*, 2014.

Zhang, L., Li, J., and Lu, H. (2016). Saliency detection via extreme learning machine. *Neurocomputing*, 218, 103–112.

Zhou, Z.-H. and Yu, Y. (2005). Ensembling local learners throughmultimodal perturbation. *IEEE Transactions on Systems, Man, and Cybernetics, Part B (Cybernetics)*, 35(4), 725–735.